\documentclass[11pt,twocolumn,letterpaper]{article}
\usepackage[margin=.9in,vmargin=.85in,columnsep=.28in]{geometry}
\usepackage{times,microtype,booktabs}
\usepackage{tikz}
\usetikzlibrary{arrows.meta,shapes.geometric}
\usepackage[round,authoryear]{natbib}

\usepackage[hidelinks]{hyperref}
\hypersetup{pdftitle={Online Learning for LLMs: A Systems Study},pdfauthor={Jesper Derehag}}

\title{\bfseries Online Learning for LLMs: A Systems Study}
\author{Jesper Derehag\\MidBrain\\\href{mailto:jesper@midbrain.ai}{\texttt{jesper@midbrain.ai}}}
\date{}
\begin{document}
\maketitle

\begin{abstract}
Online learning can eliminate a second forward pass when a language model trains on a response it has just served. In a matched comparison on NVIDIA A10G GPUs, this reduced end-to-end processing time by 33.2\% for 16-token responses and 32.5\% for 64-token responses. We measured the resources required to serve six recorded SQuAD interactions and update a Qwen2.5-1.5B adapter. The offline trainer read the raw interaction text, repeated the sequential forward computation, checkpointed the update and returned the adapter. The online inference server retained activations through decoding and applied the same update in memory. Across six 64-token responses, retention added 2.24 seconds of serving work and eliminated 34.07 seconds of replay, with bit-identical adapter updates. The principal tradeoff was GPU memory during inference: peak allocation on the serving host increased from 3.04 to 16.06\,GiB.
\end{abstract}

\section{Introduction}
\begin{figure*}[t]
\centering
\begin{tikzpicture}[font=\small,>=Stealth,
  op/.style={draw,rounded corners=2pt,fill=white,inner sep=4pt,align=center},
  wire/.style={->,semithick}, zone/.style={draw=black!45,fill=black!3,rounded corners},
  title/.style={anchor=west,font=\small\bfseries},
  tag/.style={font=\footnotesize,fill=white,inner sep=1pt}]
  \node[title] at (0,0) {(a) Offline};
  \draw[zone] (0,-.26) rectangle (7.6,-1.52);
  \node[anchor=west,font=\footnotesize] at (.08,-.48) {Inference server};
  \node[op] (s) at (1.55,-1) {Serve interaction};
  \node[op] (v) at (5.95,-1) {Installed adapter};
  \draw[wire] (v) -- node[tag,above]{weights} (s);
  \draw[zone] (2.55,-1.94) rectangle (7.6,-3.25);
  \node[anchor=west,font=\footnotesize] at (2.65,-2.16) {Offline training pipeline};
  \node[op] (b) at (3.92,-2.75) {Read \& tokenize};
  \node[op] (t) at (6.47,-2.75) {Train adapter};
  \node[draw,cylinder,shape border rotate=90,aspect=.3,align=center,inner sep=3pt]
    (log) at (1.15,-2.75) {Raw text\\log};
  \draw[wire] (s.south) -- ++(0,-.42) -| (log.north);
  \draw[wire] (log) -- (b); \draw[wire] (b) -- (t);
  \draw[wire] (t.north) -- node[tag,right]{publish} (t.north |- v.south);
  \begin{scope}[xshift=8cm]
  \node[title] at (0,0) {(b) Online};
  \draw[zone] (0,-.26) rectangle (7.6,-3.25);
  \node[anchor=west,font=\footnotesize] at (.08,-.48) {Inference server};
  \node[op] (p) at (1.05,-1.13) {Prefill};
  \node[op] (g) at (3.62,-1.13) {Generate};
  \node[op] (u) at (6.32,-1.13) {Update};
  \node[op] (n) at (6.32,-2.75) {Next request};
  \draw[wire] (p) -- (g); \draw[wire] (g) -- (u);
  \draw[wire,dashed] (p.south) -- ++(0,-.75) -|
    ([xshift=-4pt]u.south);
  \draw[wire,dashed] (g.south) -- (g.south |- 0,-1.88);
  \node[tag] at (3.63,-2.08) {prefill and decode activations};
  \draw[wire] ([xshift=4pt]u.south) -- node[tag,right]{resident adapter} ([xshift=4pt]n.north);
  \end{scope}
\end{tikzpicture}
\caption{Two paths from a served response to the next adapter. The offline trainer reconstructs the interaction from raw text; the online inference server retains activations through decoding.}\label{fig:paths}
\end{figure*}
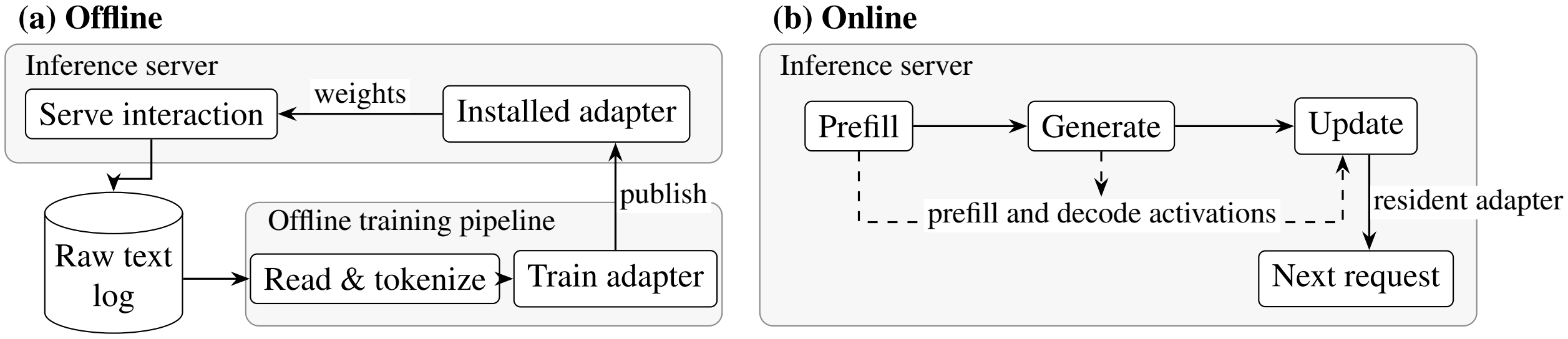

An online learner updates its model as observations arrive, so that what it learns from one example can affect its prediction on the next \citep{bottou1998}. A language-model assistant, for instance, might update a user's adapter after the user corrects an answer. An offline training pipeline reading that exchange from a text log must tokenize it and process the prompt again, even though the inference server processed the same prompt when it generated the answer (Figure~\ref{fig:paths}).

During training, the tokens of a recorded example are known in advance and the trainer can process them together. During inference, the server knows the prompt but must generate each response token before it can compute the next one. It keeps the attention keys and values for each unfinished request and can batch its next decoding step with those of other users, even when their adapters differ \citep{punica}. An inference server that also trains a user's adapter can retain the intermediate activations from prefill for backward, at the cost of keeping them in memory alongside the caches used for generation.

We assume that a learning rule and its targets have been chosen and ask what it costs to execute the update. Whether the update teaches the assistant the desired behavior is outside the scope of this paper. We address two questions:

\noindent\textbf{RQ1.} Does updating the model online save computation compared with an offline training pipeline?\\
\textbf{RQ2.} What additional resources does the inference server need when it also trains online?

For RQ1 we follow a raw-text request until the updated adapter is available for the next request. For RQ2 we measure GPU memory and time during serving, comparing a server that retains activations for a response update with one that releases them and later replays the same computation.

\section{Related work}
Dynamic evaluation adapts a language model to recently observed text \citep{krause}, while Temp-LoRA trains a temporary adapter on recent chunks during long-text generation \citep{temp}. FlexLLM shares accelerators between inference and fine-tuning even when their examples differ \citep{flex}. Alchemist retains activations from served prompts and measures training throughput during serving-idle time \citep{alchemist}. CoLLM trains a second adapter copy and switches the active snapshot for serving \citep{collm}.

Our Doc-to-LoRA analysis examined adapters compiled from documents and found that a model could retain a fact yet use it to answer the wrong question about the passage \citep{derehag2026d2l}. That finding motivated studying updates from interactions, where the request and response give an example of how the information is used.

\section{System and experimental method}
\paragraph{Matched response updates.}
We use six passages and questions from SQuAD v1.1 \citep{rajpurkar2016squad}, selecting the first question in each paragraph and the six contexts closest to 4000 characters. Each request includes the instruction to answer from the passage, the complete passage, the question and an answer cue. We use recorded responses of 16 and 64 tokens. Giving both configurations the same answer tokens in the same order ensures that they predict the same response positions even after earlier updates have changed the model. The recorded token schedule is prepared before measurement and is available only on the inference server; prompt tokenization takes place on receipt of each request.

Both configurations use Qwen2.5-1.5B with a frozen BF16 backbone and 1,089,536 trainable FP32 parameters in rank-8 LoRA adapters on the query and value projections \citep{hu}. The same eager PyTorch decoder processes the prompt and then one response token at a time with a dynamic key-value (KV) cache. Both take one AdamW step per interaction at a learning rate of $10^{-4}$, with dropout disabled, using cross-entropy over all response positions. We process requests in sequence so that the next request uses the updated weights.

\paragraph{From raw text to the next adapter.}
In the offline configuration, the inference server serves without tracking gradients and sends the raw prompt and response text to the trainer over HTTP. The trainer appends the text to a UTF-8 file, flushes it to storage, reopens the record and tokenizes both fields. It then repeats the same sequential forward computation with gradient tracking enabled, runs backward, updates the adapter, checkpoints the adapter and optimizer, and returns the adapter for installation.

In the online configuration, the inference server retains differentiable prefill and decode activations, including dependencies through the KV cache. After serving the response, it applies the same loss and updates its resident adapter directly. Neither a training log nor a per-update checkpoint is written. The common endpoint is availability of the next adapter for inference; only the offline path also provides a durable checkpoint at that point.

\paragraph{Measurement and checks.}
The testbed consists of two EC2 \texttt{g5.xlarge} instances in the same availability zone, each with one A10G. Offline training uses both hosts; online training uses only the inference host. One allocated GPU-second (GPU-s) is one second of instance time on one such host, including time waiting for the CPU or a peer. We show the allocation that each configuration requires if its participating hosts are reserved for the timed interval. Elapsed time measures how long the request takes regardless of how many hosts participate.

For each response length we first run both configurations on all six interactions and verify tokenization, logits, gradients, optimizer state and installed weights after every update. We then reset the adapter and optimizer before each six-request sequence and collect four timing repetitions, alternating which configuration runs first. Wall-clock timing begins with the raw request and ends after synchronization of the resident update or installation of the returned adapter. These measurements run without a GPU profiler or state comparisons. We sum request times and logical bytes within each sequence and report means over repetitions. Phase times partition the request interval; trainer work is counted inside the request to the trainer, never added a second time. CPU time covers the inference process and the measured trainer routine. GPU memory is the maximum allocation reported by PyTorch, and file reads may be served by the page cache.

\section{Results}
As Table~\ref{tab:system} shows, online training reduced the time to serve six requests and make their updated adapters available by 33.2\% at 16 response tokens and 32.5\% at 64 tokens. The offline trainer reconstructed the serving tokens from raw text in every check. Logits, gradients, optimizer state and installed adapters matched bit for bit after all twelve checked updates.

\begin{table*}[t]
\centering\small
\caption{Elapsed time and resource use for offline and online training, by response length.}\label{tab:system}
\begin{tabular}{@{}lrrrr@{}}
\toprule
 & \multicolumn{2}{c}{Offline} & \multicolumn{2}{c}{Online}\\
\cmidrule(lr){2-3}\cmidrule(l){4-5}
 & 16 & 64 & 16 & 64\\
\midrule
\textbf{Total elapsed time (s)} & \textbf{29.26} & \textbf{114.43} & \textbf{19.54} & \textbf{77.25}\\
Change relative to offline (\%) & N/A & N/A & $-33.2$ & $-32.5$\\
\quad Serve prompt and response (s) & 7.90 & 30.07 & 8.37 & 32.31\\
\quad Persist, reread and tokenize record (s) & 0.49 & 0.54 & N/A & N/A\\
\quad Sequential forward replay (s) & 8.23 & 34.07 & N/A & N/A\\
\quad Backward and optimizer (s) & 11.78 & 48.38 & 11.17 & 44.94\\
\quad Checkpoint and export (s) & 0.50 & 0.53 & N/A & N/A\\
\quad Communication, installation and other overhead (s) & 0.37 & 0.83 & $<0.01$ & $<0.01$\\
\midrule
Participating hosts reserved for interval (GPU-s) & 58.52 & 228.85 & 19.54 & 77.25\\
Inference-server / trainer CPU time (s) & 8.05 / 20.86 & 30.25 / 83.39 & 19.54 / N/A & 77.27 / N/A\\
Peak inference memory during serving (GiB) & 3.03 & 3.04 & 7.89 & 16.06\\
\midrule
Raw log, written and reread (kB each) & 20.89 & 22.12 & 0 & 0\\
HTTP request / return bodies (MB) & 0.026 / 26.38 & 0.028 / 26.38 & 0 & 0\\
Adapter and optimizer checkpoints written (MB) & 79.28 & 79.28 & 0 & 0\\
\bottomrule
\end{tabular}
\end{table*}

\paragraph{Time saved by updating online (RQ1).}
Across six 64-token responses, retaining activations increased serving time from 30.07 to 32.31\,s but avoided 34.07\,s of sequential replay. Backward and the optimizer step took 48.38\,s offline and 44.94\,s online. The 32.5\% saving measures elapsed time until the updated adapter is ready for inference, including the time spent logging, checkpointing and transferring weights in the offline configuration.

During offline generation the trainer waits for the inference server; during training the inference server waits for the updated adapter. Reserving both hosts for the six 64-token requests therefore accounted for 228.85 GPU-s, compared with 77.25 GPU-s on one host online. A trainer shared across users could put some of the waiting time to other use.

\paragraph{Resources needed on the inference server (RQ2).}
At 64 tokens, peak serving memory increased from 3.04 to 16.06\,GiB. Only the adapter parameters are updated, but their gradients depend on activations throughout prefill and decoding. Offline training incurred that memory demand later on the trainer, whose peak allocation reached 16.11\,GiB. Online training needed the memory on the inference host while the request was still active. Over the same six requests, the inference process used 77.27 CPU-seconds online and 30.25 offline; the offline trainer used a further 83.39 CPU-seconds in its measured phases.

\section{Discussion and limitations}
The offline trainer in this comparison serves one user and repeats the same sequential forward pass as the inference server. This makes it possible to check that the updates agree, but an offline trainer with known answers could process response positions in parallel or batch examples across users \citep{fusion}. Concurrency also changes the cost of online training: memory held for backward may otherwise have served another request. The experiment measures neither effect under load. The two deployments also differ in durability. A service that archives transcripts or requires recovery after every update would still incur those writes online.

For procedural memory, the value of an interaction lies partly in the circumstances it records. A user who removes background from an email draft but keeps detailed directions for the recipient gives a more specific example than the instruction to keep emails short. Such a correction also illustrates a limit of the measured saving: the retained activations cover the original draft, whereas predicting tokens after an edited prefix requires new computation. If feedback arrives after the activations have been released, training may require a full replay.

\section{Conclusion}
For the matched response updates studied here, keeping the adapter on the inference server reduced the time from a raw request to the next usable adapter by about one third. Retaining the serving activations avoided a second forward pass at the cost of higher memory demand during inference. For a service that can accommodate those activations, the result suggests a lower cost of acquiring an update; any further saving from omitting a learned instruction on later requests depends on the behavior that update produces.

\paragraph{Future work.}
We plan to study context distillation as a learning signal for procedural memory, including whether an adapter learns when a procedure applies and retains that behavior as the user's instructions change.

\bibliographystyle{plainnat}
\bibliography{references,systems_refs,learning_refs}
\end{document}